\documentclass[journal]{IEEEtran}
\usepackage[utf8]{inputenc}
\usepackage{hyperref}
\usepackage{amsmath}
\usepackage{graphicx}		

\usepackage{svg}
\usepackage{academicons}
\newcommand{\orcidicon}{\includegraphics[width=8pt]{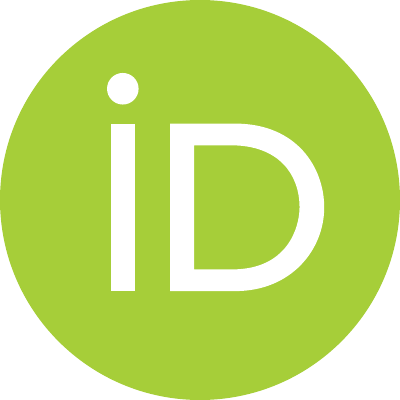}}

\usepackage{arydshln}
\usepackage[normalem]{ulem}
\usepackage[linesnumbered,ruled,vlined]{algorithm2e}
\usepackage{amssymb}
\usepackage{xcolor}
\usepackage{algpseudocode}
\usepackage{booktabs}
\usepackage{multirow}
\usepackage{graphicx}
\usepackage{colortbl}
\usepackage{makecell}
\usepackage{enumitem}
\usepackage{hyperref}
\usepackage{listings}
\usepackage[T1]{fontenc}
\usepackage[utf8]{inputenc}

\usepackage[linesnumbered,ruled,vlined]{algorithm2e}
\usepackage{amssymb}
\usepackage{xcolor}
\usepackage{algpseudocode}
\usepackage{booktabs}
\usepackage{multirow}
\usepackage{graphicx}
\usepackage{colortbl}
\usepackage{makecell}
\usepackage{enumitem}
\usepackage{hyperref}
\usepackage{listings}
\usepackage[T1]{fontenc}
\usepackage[utf8]{inputenc}
\usepackage{rotating}
\usepackage{pifont}

\usepackage{listings}
\usepackage{bm}

\usepackage{hyperref}
\usepackage{amsmath}
\usepackage{graphicx}		

\usepackage{svg}
\usepackage{academicons}

\usepackage{arydshln}
\usepackage[normalem]{ulem}

\ifCLASSINFOpdf
\else
\fi
\begin{document}
%

\title{${\mathcal N}^{\mathcal Segment}$: Label-specific Deformations \\for Remote Sensing Image Segmentation \vspace{-0.0cm}}

%
%
%



\author{\IEEEauthorblockN{Yechan Kim\textsuperscript{\href{https://orcid.org/0000-0002-2438-3590}{\orcidicon{}}}, \textit{Member}, \textit{IEEE}}, 
\IEEEauthorblockN{DongHo Yoon\textsuperscript{\href{https://orcid.org/0009-0006-1514-7293}{\orcidicon{}}}}, 
\IEEEauthorblockN{SooYeon Kim\textsuperscript{\href{https://orcid.org/0009-0005-1474-6828}{\orcidicon{}}}}, and 
\IEEEauthorblockN{Moongu Jeon\textsuperscript{\href{https://orcid.org/0000-0002-2775-7789}{\orcidicon{}}}, \textit{Senior Member}, \textit{IEEE}}
\vspace{-0.8cm}
\thanks{Received \textcolor{black}{1 July 2025}; revised \textcolor{black}{27 July 2025}; accepted \textcolor{black}{00 00000000 2025}; Date of publication \textcolor{black}{00 00000000 2025}; date of current version \textcolor{black}{00 00000000 2025}. \textcolor{black}{This work was supported by the Agency For Defense Development Grant funded by the Korean Government (UI220066WD). We appreciate the high-performance GPU computing support of HPC-AI Open Infrastructure via GIST SCENT.} (Yechan Kim and DongHo Yoon contributed equally to this work.) (\textit{Corresponding authors}: \textit{Moongu Jeon; Yechan Kim.})}
\thanks{The authors are with Machine Learning and Vision Laboratory (MLV), Gwangju Institute of Science and Technology (GIST), Gwangju 61005, South Korea (e-mail: 
    \href{mailto:yechankim@gm.gist.ac.kr}{yechankim@gm.gist.ac.kr}, 
    \href{mailto:gidong76@gm.gist.ac.kr}{gidong76@gm.gist.ac.kr},
    \href{mailto:bluesooyeon@gm.gist.ac.kr}{bluesooyeon@gm.gist.ac.kr},
    \href{mailto:mgjeon@gist.ac.kr}{mgjeon@gist.ac.kr}).}
\thanks{This article has supplementary downloadable material available at \url{https://doi.org/10.1109/LGRS.2025.xxxxxxxx}, provided by the authors.}
\thanks{Digital Object Identifier 10.1109/LGRS.2025.xxxxxxx}}

%
%

\markboth{}%
{Shell \MakeLowercase{\textit{et al.}}: Bare Demo of IEEEtran.cls for Journals}
%



\maketitle

\begin{abstract}
Labeling errors in remote sensing (RS) image segmentation datasets often remain implicit and subtle due to ambiguous class boundaries, mixed pixels, shadows, complex terrain features, and subjective annotator bias.
    Furthermore, the scarcity of annotated RS data due to the high cost of labeling complicates training noise-robust models.
While sophisticated mechanisms such as label selection or noise correction might address the issue mentioned above, they tend to increase training time and add implementation complexity.
    In this paper, we propose \textbf{\textit{NSegment}}—a simple yet effective data augmentation solution to mitigate this issue. 
        Unlike traditional methods, it applies elastic transformations only to segmentation labels, varying deformation intensity per sample in each training epoch to address annotation inconsistencies.
Experimental results demonstrate that our approach improves the performance of RS image segmentation over various state-of-the-art models.
\end{abstract}


%
\IEEEpeerreviewmaketitle

\section{Introduction}
%
%
%
%
\IEEEPARstart{S}{egmentation} tasks in remote sensing (RS) imagery play a key role in numerous applications, including land cover classification, urban planning, and military surveillance.
    However, training noise-robust RS segmentation models remains a challenge due to inherent labeling errors and annotation inconsistencies \cite{wang2021loveda}.
        These arise from ambiguous class boundaries, mixed pixels, shadows, complex terrain features, and subjective annotator biases.
    In addition, the scarcity of annotated RS data, caused by the high labeling cost, further complicates model training, limiting the ability of generalization against such implicit noisy labels \cite{wang2024mtp}.

Previous studies have explored various approaches, such as label selection strategies \cite{liu2024cromss}, noise correction techniques \cite{liu2024aio2}, and uncertainty modeling \cite{landgraf2024uncertainty}.
    While these methods might improve model robustness, they often come at the cost of increased computational complexity and prolonged training time, particularly due to extensive hyperparameter tuning.
        Additionally, although they have addressed `severe' label noise, they did not explicitly account for `subtle' (or implicit) label noise\footnote{This work treats implicit, inherent, and subtle label noise as equivalent. Note that we focus on a form of label noise that is inherently subtle and not overtly erroneous, thus challenging to identify even by human experts.} that may emerge from ambiguity or bias.

Subtle annotation noise may seem negligible on a per-instance basis, but when such errors recur inconsistently throughout a dataset, their cumulative effect can impair model generalization.
        To address this issue, we aim to develop a lightweight yet effective strategy for handling such inherent annotation noise in existing RS segmentation datasets.
    One straightforward approach is data augmentation, which increases the diversity of training samples, making it more resilient to potential annotation noise \cite{song2024exploring, kim2025nbbox}.
    


This letter presents a simple yet efficient data augmentation method, namely Noise injection into Segmentation labels (\textbf{\textit{NSegment}}), for RS image segmentation.
        Unlike conventional augmentation techniques that identically apply global transformations to both images and labels, our method selectively perturbs only the segmentation labels while preserving the original image content.
            Moreover, our approach varies the deformation intensity per sample in each training epoch, ensuring diverse label-level augmentations.
    This process allows models to learn from potentially inconsistent annotations while maintaining global structural integrity.

We evaluate \textbf{\textit{NSegment}} across multiple state-of-the-art models and benchmark datasets, including ISPRS Vaihingen \cite{vaihingen}, ISPRS Potsdam \cite{potsdam}, and LoveDA \cite{wang2021loveda}. 
    Experimental results demonstrate that our method significantly improves RS image segmentation performance by enhancing label robustness, making it a practical solution for real-world RS applications.

\vspace{-0.2cm}
\section{Methodology}

This section introduces a simple yet effective data augmentation solution, namely \textbf{\textit{NSegment}} for RS image segmentation.
\vspace{-0.8cm}

\begin{figure*}[t!]
    \centering
    \includegraphics[width=17cm]{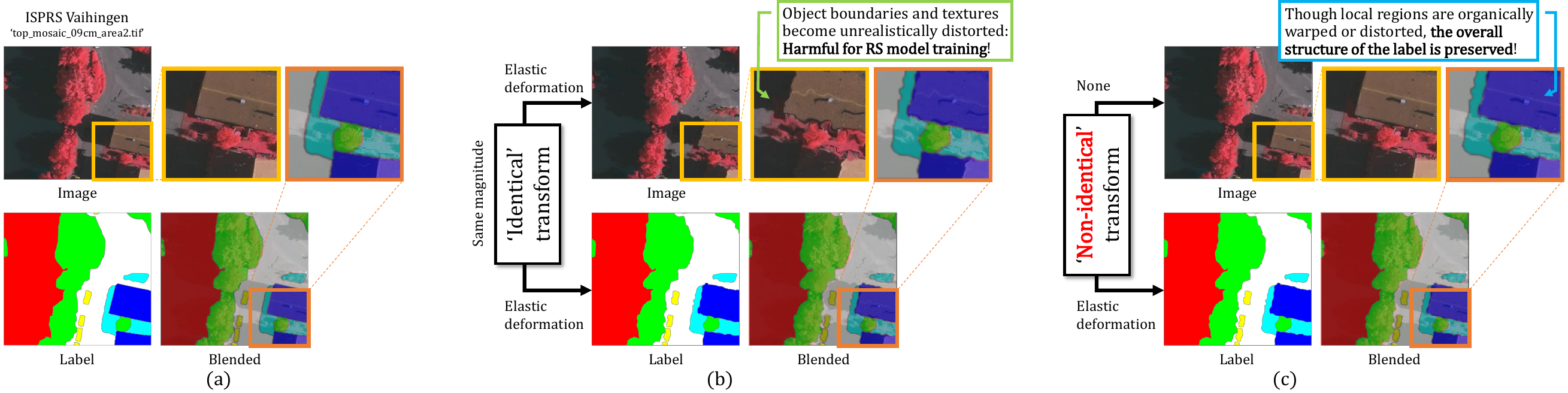} \vspace{-0.3cm}
    \caption{{Illustration of the proposed data augmentation method named \textbf{\textit{NSegment}} for RS image segmentation. (a) Original image-label pair. (b) Typical use of elastic deformation for data augmentation. (c) Proposed approach. Best viewed on a colored screen with zoom.}}
    \label{fig1}
   \vspace{-0.3cm}
\end{figure*}

\subsection{Background}
    Elastic deformation is a technique used to introduce smooth, non-rigid distortions to an image while preserving its continuous structure, typically without introducing discontinuities or artifacts.
        This concept originates from material science, describing reversible shape changes that occur under applied forces \cite{rivlin1948large}. 
    Elastic deformation is also widely used in various computer vision tasks.
        For example, Simard et al. \cite{simard2003best} pioneered its use for handwritten character recognition, where random elastic distortions were applied to training samples to improve models' generalization.
    Large Deformation Diffeomorphic Metric Mapping \cite{glaunes2008large} is one of the most well-known approaches based on elastic transformation in image registration and shape classification.
    Recently, elastic deformation has been adopted in medical image recognition to increase data diversity, helping models become more invariant to anatomical variations 
    \cite{islam2024systematic}.
        Building upon this research line, we extend the concept of elastic deformation by varying the transformation intensity per sample for RS image segmentation.
    Besides, unlike typical augmentations that apply `identical' transformations to images and labels, we introduce a `decoupled' transformation strategy inspired by NBBOX \cite{kim2025nbbox}, which only perturbs localization annotations, leaving the corresponding image unchanged. 
        Unlike NBBOX, applying object-wise linear geometric perturbation, our \textbf{\textit{NSegment}} employs non-rigid, spatially smooth label deformations for RS segmentation\footnote{Further differences between NBBOX and our \textbf{\textit{NSegment}} are described in Section A of the Supplementary Material.}.


\vspace{-0.2cm}
\begin{algorithm}[t!]
\label{algo1}
\caption{Procedure of \textbf{\textit{NSegment}}.}
\SetKwInOut{Initialization}{Parameters}
\KwIn{Input image $\mathcal{I}$, Corresponding segmentation labels $\mathcal{L} = ({S}_j)_{j=1}^{C}$, Set of $(\alpha, \sigma)$ combinations $\mathbf{\Omega} = \{(\alpha_k, \sigma_k)\}_{k=1}^{K}$, Transform probability $p$}
    \If{$\textit{rand}() > p$} {
        \textbf{return} $\mathcal{L}$\tcp*{\small\textnormal{No update labels}} 
    }
    $w, h \gets \mathcal{I}.\textit{shape}$\;
    Initialize ${S}_j^{*}$ for each category index $j$ as zero-filled matrix of shape $w\times h$\;
    $\alpha, \sigma \gets \textit{rand\_choice}(\mathbf{\Omega})$\;
    $d\mathcal{X} \gets G_{\sigma} \circledast \phi (\alpha (2 \cdot \textit{rand}(w, h) - 1))$\;
    $d\mathcal{Y} \gets G_{\sigma} \circledast \phi (\alpha (2 \cdot \textit{rand}(w, h) - 1))$\;
    \For{$j \gets 1$ \textbf{to} $C$}{
        Set ${S}_j^{*}[\varphi(x + d\mathcal{X}[x, y]), \varphi(y + d\mathcal{Y}[x, y])]$ to ${S}_j[x, y] \text{ for each pixel } (x, y) \text{ in segment } {S}_j$\;
    }
$\mathcal{L}^* \gets (S_j^{*})_{j=1}^{C}$\;
\textbf{return} $\mathcal{L}^*$\tcp*{\small\textnormal{Update labels}} 
\end{algorithm}

\vspace{-0.2cm}
\subsection{Overview of the proposed method}

\begin{figure}[!t]
    \centering
    \includegraphics[width=7.4cm]{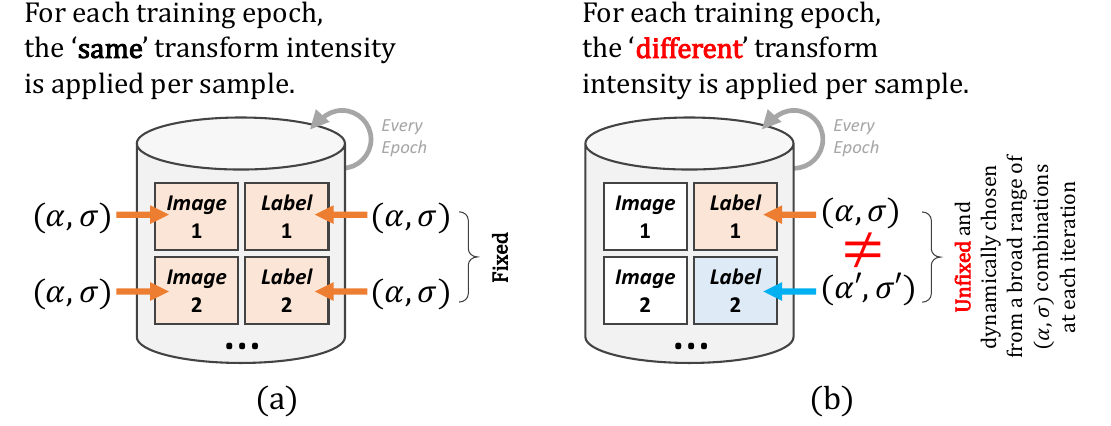} \vspace{-0.4cm}
    \caption{Detailed comparison of (a) typical elastic deformation and (b) proposed approach for each training epoch in terms of magnitude diversity.}
    \label{fig2}
    \vspace{-0.3cm}
\end{figure}

The detail of \textbf{\textit{NSegment}} is shown in Algorithm \ref{algo1} (for one epoch/sample).   
        Our method is a lightweight yet effective label-level data augmentation strategy designed to improve the robustness of RS segmentation models.
It aims to mimic potential inconsistencies and inherent noise in annotations while preserving global structures.
    The core part of \textbf{\textit{NSegment}} is to introduce controlled elastic deformations only to segmentation labels by applying spatially coherent displacement fields as shown in Fig. \ref{fig1}.
Besides, as indicated in Fig. \ref{fig2}, by randomly sampling degrees of deformation strength (magnitude) and smoothness from a wide parameter range, it avoids the need for exhaustively tuning the hyperparameters while yielding consistent performance gains, as it allows the model to learn diverse variants of labels (i.e., label-level regularization). 

To explain our algorithm, we provide an example illustrating how the label of a single image $\mathcal{I}$ is transformed.
    Let $\mathcal{I}$ be one of the training images. $S_{j}$ is the segmentation label mask of the $j$th class for the input image. $C$ represents the number of classes for semantic segmentation.
        The goal is to generate a new label $\mathcal{L}^*=(S_j^{*})_{j=1}^{C}$ from $\mathcal{L} =(S_j^{})_{j=1}^{C}$ as:

\begin{itemize}
    \item (Step 1) Initialize segmentation masks $S_j^{*}$ as zero-filled matrices of shape $w \times h$ for each category $j$, where $w$ and $h$ are width and height of $\mathcal{I}$.
    \item (Step 2) With probability $(1-p)$, retain the original mask and skip label transformation. Otherwise, randomly select $(\alpha, \sigma)$ from $\mathbf{\Omega}=\{(\alpha_k, \sigma_k)\}_{k=1}^{K}$ and go to (Step 3). $p$ and $\mathbf{\Omega}$ are to be tuned and $\alpha$ and $\sigma$ are defined in (Step 3).
    \item (Step 3-1) Generate random displacement fields of $x$- and $y$-axis and smooth them with the Gaussian filter as in lines 6-7 of Algorithm \ref{algo1}, where $\alpha$ determines the `magnitude of displacement', $\circledast$ represents the convolution operation, $\textit{rand}(w, h)$ is a matrix of shape $w \times h$ with elements sampled uniformly from $[0, 1]$, $\phi$ is zero-padding to the input such that the output of $\circledast$ preserves the same spatial dimensions as the input, and $G_{\sigma}$ means the 2D Gaussian kernel as follows:
        \begin{equation}\label{eq1}
        \begin{split}
            & \textstyle G_{\sigma}(m, n) = \frac{1}{2 \pi \sigma^2} e^{- \frac{m^2+n^2}{2 \sigma^2}},  \text{ }m,n\in[-r, r].
        \end{split}
        \end{equation}
        Here, $\sigma$ is the Gaussian filter's standard deviation, controlling the deformation's spatial `smoothness', and $r$ is the half-size of the filter. 
            Based on Eq. (\ref{eq1}), the $x$-axis smooth random displacement field $d\mathcal{X}$ is calculated as:
                \begin{equation}\label{eq2}
                \begin{split}
                    & \textstyle d\mathcal{X}[x, y] = \sum_{m,n} G_{\sigma}(m, n) \cdot \{ \alpha (2 \cdot \textit{rand}() -1) \}, \\ 
                \end{split} \vspace{-0.0cm}
                \end{equation}
                where \textit{rand}() is a random generator for $R \in [0,1]$. 
                Similarly, $d\mathcal{Y}$ can be derived analogously to $d\mathcal{X}$.
    \item (Step 3-2) Apply elastic deformation to each $j$th segmentation $S_j^{}$. 
            For each pixel $(x, y)$ in $S_j^{}$, the deformed position is computed as $(\varphi(x + d\mathcal{X}[x, y]), \varphi(y + d\mathcal{Y}[x, y]))$, where $\varphi(\cdot)$ ensures index validity to prevent out-of-bound errors. Finally, $S_j^{*}$ for each $j$ is obtained as in line 9.
\end{itemize}


\begin{table}[t!]
\vspace{-0.3cm} 
\centering
\caption{{\label{tab:table-name} ($\alpha$, $\sigma$) Parameter study of image- and label-level elastic deformations with PSPNet on the ISPRS Vaihingen dataset}} \vspace{-0.2cm}
\resizebox{8cm}{!}{%
\begin{tabular}{@{}cc|ccccc|c@{}}
\toprule
\multicolumn{2}{c|}{(Baseline mIoU: 77.41)} & $\alpha$=1 & $\alpha$=15 & $\alpha$=30 & $\alpha$=50 & $\alpha$=100 & \begin{tabular}[c]{@{}c@{}}All $(\alpha, \sigma)$ combinations \\ except for $\sigma$=1\end{tabular} \\ \midrule
\multicolumn{1}{c|}{\multirow{4}{*}{Image-level only}} & $\sigma$=1 & 76.87& 65.37& 54.65& 32.23& 14.40& \multirow{4}{*}{70.69} \\
\multicolumn{1}{c|}{} & $\sigma$=3 &  76.92&  76.66&  75.72&  74.75&  70.67&  \\
\multicolumn{1}{c|}{} & $\sigma$=5 &  \underline{77.01}&  76.58&  76.26&  75.93&  74.30&  \\
\multicolumn{1}{c|}{} & $\sigma$=10 &  76.99&  \underline{76.75}&  \underline{76.49}&  \underline{76.68}&  \underline{76.40}&  \\ \midrule
\multicolumn{1}{c|}{\multirow{4}{*}{Label-level only}} & $\sigma$=1 &  77.27&  77.55&  77.07&  74.26&  63.51& \multirow{4}{*}{\textbf{77.75}} \\
\multicolumn{1}{c|}{} & $\sigma$=3 &  \underline{77.57}&  77.50&  77.51&  77.25&  75.99&  \\
\multicolumn{1}{c|}{} & $\sigma$=5 &  77.30&  \underline{77.70}&  77.41& 77.24 &  77.54 &  \\
\multicolumn{1}{c|}{} & $\sigma$=10 &  76.49&  77.34&  \underline{77.67}&  \underline{77.55}& \underline{77.66} &  \\ 
\bottomrule 
\multicolumn{8}{c}{(For both image-level and label-level metrics, the highest values per column are \underline{underlined},} \\
\multicolumn{8}{c}{with the overall maximum emphasized in \textbf{bold}.)}
\end{tabular}%
} \vspace{-0.1cm}
\label{tab1}
\end{table}

\begin{table}[t!]
\vspace{-0.1cm} 
\centering
\caption{{\label{tab:table-name} Comparison of three elastic deformation types under random ($\alpha$, $\sigma$) pairs with PSPNet on the ISPRS Vaihingen dataset}} \vspace{-0.2cm}
\resizebox{7.1cm}{!}{%
\begin{tabular}{@{}l|ccc|c@{}}
\toprule
 \multicolumn{1}{c|}{\multirow{2.5}{*}{\begin{tabular}[c]{@{}c@{}}For each case, the model is trained\\from the same weight initialization\end{tabular}}} & \multicolumn{3}{c|}{Elastic deformation} & \multicolumn{1}{c}{\multirow{2.5}{*}{\begin{tabular}[c]{@{}c@{}}Test \\mIoU\end{tabular}}} \\ \cmidrule(lr){2-4}
 & Image & Label & Identical? &  \\ \midrule
Baseline (No augmentation used)& -& -& -& 77.41\\
(a) Normal (Identical image-label transform) & \checkmark & \checkmark & Yes & 67.58\\
(b) Transform only for images & \checkmark  &  & No & 70.69\\
\cellcolor[HTML]{FFCE93}(c) Transform only for labels (Ours) &\cellcolor[HTML]{FFCE93}  & \cellcolor[HTML]{FFCE93}\checkmark & \cellcolor[HTML]{FFCE93}No & \cellcolor[HTML]{FFCE93}\textbf{77.75}\\ \bottomrule
\end{tabular}
} \vspace{-0.2cm}
\label{tab2}
\end{table}

\vspace{-0.2cm}
\section{Experiments}
\label{sec3}
\vspace{-0.0cm}
This section presents the experimental results of \textbf{\textit{NSegment}}, demonstrating its effectiveness in RS image segmentation.

\vspace{-0.2cm}
\subsection{Experimental setup}

        Unlike conventional studies that focus on explicit annotation errors, we assume 1) no obviously incorrect labels exist in the datasets, and 2) all data are affected by inherent and implicit label noise, as true labels are indeed unobservable \cite{chen2021noise}.
    The Gaussian kernel size for each $\sigma$ is automatically determined as $2 \cdot \left\lfloor 3 \sigma \right\rceil + 1$, following OpenCV's setting.
        Note that in our algorithm, all for-loops can be parallelizable, thus training time differences with and without our method are negligible\footnote{Python code implementation and experimental environments are presented in Sections B and C of the supplementary material.}.
    

    

\vspace{-0.2cm}
\subsection{Datasets}
We use three commonly used datasets named ISPRS Vaihingen \cite{vaihingen}, ISPRS Potsdam \cite{potsdam}, and LoveDA \cite{wang2021loveda} for RS image segmentation\footnote{Details about each dataset, including spatial resolution, semantic classes, and train-test splits, are in Section D of the supplementary material.}.
    Specifically, the Vaihingen and Potsdam datasets, released by the ISPRS, consist of high-resolution true orthophoto tiles captured from urban areas using airborne sensors, making them standard benchmarks for urban semantic segmentation.
        In contrast, the LoveDA dataset comprises satellite images covering both urban and rural environments, enabling evaluations across diverse land-use types.

\vspace{-0.2cm}
\subsection{Investigating elastic deformation in RS image segmentation}
This subsection investigates the effects of applying elastic transformation separately to RS images and their corresponding labels. 
    For segmentation masks, Algorithm \ref{algo1} can be applied directly. 
        In the case of images, however, the same algorithm should be modified by simply switching the transformation target from the label to the image. 
    Through a comparative analysis, we highlight the necessity of our original proposed \textbf{\textit{NSegment}} framework that solely perturbates segmentation masks during training. 
        In particular, we conduct an extensive exploration\footnote{Section E of the supplementary material details the behavioral characteristics of $\alpha$ and $\sigma$ with extreme cases, to aid reader understanding.} across various combinations of the transformation parameters, $\alpha$ and $\sigma$, to identify optimal settings, with fixing $p$ as 1.
    For this, we use PSPNet, adopting ResNet-18 with the Vaihingen dataset.
        For the sake of experimental objectivity, each experiment is conducted for 80,000 iterations under the same weight initialization and random seed.

\begin{figure}[!t]
    \centering
    \includegraphics[height=4.2cm]{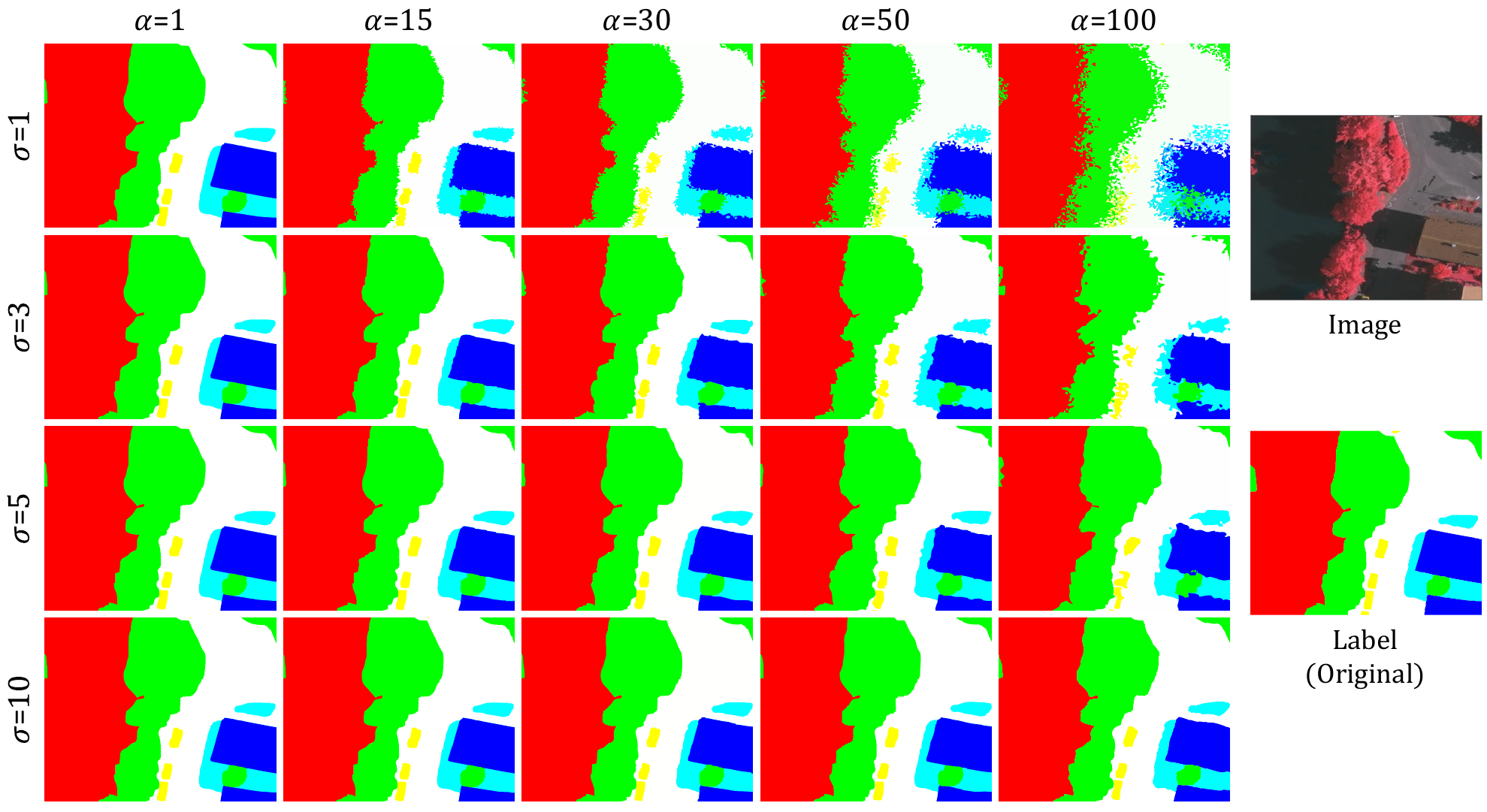} \vspace{-0.3cm}
    \caption{Illustration of elastic transform effects on the segmentation label across various $(\alpha, \sigma)$ settings. Best viewed on a colored screen with zoom.}
    \label{fig3}
    \vspace{-0.3cm}
\end{figure}

\begin{table*}[!t]
\vspace{-0.3cm} 
\centering
\caption{{\label{tab:table-name} Impact of \textbf{\textit{NSegment}} on training state-of-the-art segmentation models on the ISPRS Vaihingen, Potsdam, and LoveDA datasets}} \vspace{-0.2cm}
\resizebox{16.5cm}{!}{%
\begin{tabular}{@{}c|c|ccccccccc@{}}
\toprule
\multirow{2.5}{*}{Data} & Model & PSPNet \cite{zhao2017pyramid} & DeepLabV3+ \cite{chen2018encoder} & DANet \cite{fu2019dual} & OCRNet \cite{yuan2020object} & SegFormer \cite{xie2021segformer} & DDRNet \cite{pan2022deep} & DOCNet \cite{ma2023docnet} & RS$^3$Mamba \cite{ma2024rs} &  LOGCAN++ \cite{ma2025logcan++} \\ \cmidrule(l){2-11} 
 & Publication & (CVPR'17) & (ECCV'18) & (CVPR'19) & (ECCV'20) & (NeurIPS'21) & (TITS'22) & (GRSL'23) & (GRSL'24) & (TGRS'25) \\ \midrule
\multirow{2.5}{*}{\begin{tabular}[c]{@{}c@{}}ISPRS\\ Vaihingen\end{tabular}} & Baseline & 77.92 (77.63$\pm$0.22) & 77.98 (77.53$\pm$0.30) & 80.20 (79.59$\pm$0.61) & 77.99 (76.96$\pm$0.84) & 79.25 (79.05$\pm$0.11) & 74.69 (74.32$\pm$0.47) & 81.04 (80.80$\pm$0.32) & 80.87 (80.48$\pm$0.40) & 81.05 (80.97$\pm$0.08) \\ \cmidrule(l){2-11} 
 & \cellcolor[HTML]{FFCE93}+ Ours & \cellcolor[HTML]{FFCE93}\textbf{78.60} (\textbf{77.77}$\pm$0.58)  & \cellcolor[HTML]{FFCE93}\textbf{79.02} (\textbf{78.26}$\pm$0.63) & \cellcolor[HTML]{FFCE93}\textbf{80.24} (\textbf{79.71}$\pm$0.24) & \cellcolor[HTML]{FFCE93}\textbf{78.25} (\textbf{77.76}$\pm$0.45) & \cellcolor[HTML]{FFCE93}\textbf{79.45} (\textbf{79.27}$\pm$0.16) & \cellcolor[HTML]{FFCE93}\textbf{75.27} (\textbf{74.76}$\pm$0.43) & \cellcolor[HTML]{FFCE93}\textbf{81.58} (\textbf{81.41}$\pm$0.15) & \cellcolor[HTML]{FFCE93}\textbf{81.21} (\textbf{80.81}$\pm$0.34) & \cellcolor[HTML]{FFCE93}\textbf{81.19} (\textbf{81.04}$\pm$0.14) \\ \midrule
\multirow{2.5}{*}{\begin{tabular}[c]{@{}c@{}}ISPRS\\ Potsdam\end{tabular}} & Baseline & 83.18 (\textbf{83.06}$\pm$0.07) & 83.18 (82.95$\pm$0.10) & 84.80 (84.47$\pm$0.24) & 83.37 (83.08$\pm$0.18) & 84.46 (84.42$\pm$0.03) & 81.55 (81.34$\pm$0.18) & 85.67 (85.61$\pm$0.05) & 85.26 (84.78$\pm$0.28)  & 86.13 (86.07$\pm$0.06)  \\ \cmidrule(l){2-11} 
 & \cellcolor[HTML]{FFCE93}+ Ours & \cellcolor[HTML]{FFCE93}\textbf{83.19} (83.03$\pm$0.09) & \cellcolor[HTML]{FFCE93}\textbf{83.27} (\textbf{83.16}$\pm$0.11) & \cellcolor[HTML]{FFCE93}\textbf{84.87} (\textbf{84.51}$\pm$0.23) & \cellcolor[HTML]{FFCE93}\textbf{83.40} (\textbf{83.23}$\pm$0.14) & \cellcolor[HTML]{FFCE93}\textbf{84.50} (\textbf{84.43}$\pm$0.05) & \cellcolor[HTML]{FFCE93}\textbf{81.56} (\textbf{81.42}$\pm$0.13) & \cellcolor[HTML]{FFCE93}\textbf{85.74} (\textbf{85.69}$\pm$0.05) & \cellcolor[HTML]{FFCE93}\textbf{85.48} (\textbf{85.04}$\pm$0.36)  & \cellcolor[HTML]{FFCE93}\textbf{86.27} (\textbf{86.11}$\pm$0.14)   \\ \midrule
 \multirow{2.5}{*}{LoveDA} & Baseline & 43.92 (43.11$\pm$0.53) &  44.33 (43.29$\pm$0.77) & 44.45 (43.35$\pm$0.78) & 46.90 (46.45$\pm$0.47) & \textbf{48.59} (47.91$\pm$0.57) &  \textbf{42.00} (40.57$\pm$1.35) & 51.45 (50.92$\pm$0.50) & 46.93 (46.42$\pm$0.59)  & 50.68 (50.59$\pm$0.13)  \\ \cmidrule(l){2-11} 
 & \cellcolor[HTML]{FFCE93}+ Ours & \cellcolor[HTML]{FFCE93}\textbf{44.35} (\textbf{43.55}$\pm$0.65)  & \cellcolor[HTML]{FFCE93}\textbf{44.42} (\textbf{43.36}$\pm$0.62) & \cellcolor[HTML]{FFCE93}\textbf{44.55} (\textbf{43.82}$\pm$0.89) & \cellcolor[HTML]{FFCE93}\textbf{47.31} (\textbf{46.49}$\pm$0.48) & \cellcolor[HTML]{FFCE93}48.11 (\textbf{47.92}$\pm$0.27) & \cellcolor[HTML]{FFCE93}41.68 (\textbf{40.60}$\pm$0.98) & \cellcolor[HTML]{FFCE93}\textbf{51.59} (\textbf{51.12}$\pm$0.42) & \cellcolor[HTML]{FFCE93}\textbf{47.45} (\textbf{47.04}$\pm$0.35)  & \cellcolor[HTML]{FFCE93}\textbf{51.16} (\textbf{50.77}$\pm$0.34)  \\  \bottomrule
\end{tabular}
} \vspace{-0.3cm}
\label{tab3}
\end{table*}

\begin{table}[ht!]
\vspace{-0.1cm} 
\centering
\caption{{\label{tab:table-name} Impact of combining \textbf{\textit{NSegment}} and existing augmentation techniques with DeepLabV3+ on the ISPRS Vaihingen, Potsdam, and LoveDA datasets}} \vspace{-0.2cm}
\resizebox{8cm}{!}{%
\begin{tabular}{@{}l|ccc@{}}
\toprule
\multicolumn{1}{c|}{\multirow{2}{*}{\begin{tabular}[c]{@{}c@{}}Each model is trained\\from the same weight initialization\end{tabular}}} & \multicolumn{1}{c}{\multirow{2}{*}{\begin{tabular}[c]{@{}c@{}}ISPRS\\Vaihingen\end{tabular}}} & \multicolumn{1}{c}{\multirow{2}{*}{\begin{tabular}[c]{@{}c@{}}ISPRS\\Potsdam\end{tabular}}} & LoveDA \\ \\ \midrule
Baseline (B)                 & 77.98 (77.53$\pm$0.30) & 83.18 (82.95$\pm$0.10) & 44.33 (43.29$\pm$0.77) \\
B + Horizontal Flipping (HF) & 78.82 (78.62$\pm$0.19) & 84.15 (84.00$\pm$0.12) & 43.68 (43.48$\pm$0.22) \\
\cellcolor[HTML]{FFCE93}B + HF + Ours$^{\dagger}$  & \cellcolor[HTML]{FFCE93}79.09 (78.88$\pm$0.23)  & \cellcolor[HTML]{FFCE93}84.10 (84.09$\pm$0.01) & \cellcolor[HTML]{FFCE93}44.37 (43.83$\pm$0.47) \\
B + Random Resize (RR) (0.5x-2.0x) & 80.46 (80.04$\pm$0.38) & 84.52 (84.37$\pm$0.13) &41.93 (40.73$\pm$1.05) \\
\cellcolor[HTML]{FFCE93}B + RR + Ours$^{\dagger}$  & \cellcolor[HTML]{FFCE93}\textbf{80.59} (\textbf{80.36}$\pm$0.24)  & \cellcolor[HTML]{FFCE93}84.57 (84.39$\pm$0.22) & \cellcolor[HTML]{FFCE93}45.26 (45.11$\pm$0.17) \\
B + HF + RR & 80.25 (80.02$\pm$0.34) & 84.87 (84.83$\pm$0.06) & 41.33 (40.66$\pm$0.80) \\
\cellcolor[HTML]{FFCE93}B + HF + RR + Ours$^{\dagger}$  & \cellcolor[HTML]{FFCE93}80.26 (\underline{80.22}$\pm$0.06)  & \cellcolor[HTML]{FFCE93}\textbf{85.13} (\textbf{85.04}$\pm$0.09) & \cellcolor[HTML]{FFCE93}\underline{45.84} (44.65$\pm$1.03) \\
B + HF + RR + CutMix (CM) \cite{yun2019cutmix}  & 80.18 (80.08$\pm$0.09) & 84.90 (84.70$\pm$0.26)  & 45.68 (\underline{45.29}$\pm$0.49)  \\
\cellcolor[HTML]{FFCE93}B + HF + RR + CM + Ours$^{\dagger}$ & \cellcolor[HTML]{FFCE93}\underline{80.44} (80.18$\pm$0.24) & \cellcolor[HTML]{FFCE93}\underline{85.09} (\underline{84.93}$\pm$0.14)  & \cellcolor[HTML]{FFCE93}\textbf{45.99} (\textbf{45.45}$\pm$0.71) \\ \bottomrule
\multicolumn{4}{c}{(For each dataset, the highest maximum/mean mIoU values: \textbf{boldfaced} while second-best: \underline{underlined}.)} \\
\multicolumn{4}{c}{($^{\dagger}$Ours is applied right after loading the images and labels, followed by other transforms like HF and RR.)} \\
\end{tabular}
} \vspace{-0.3cm}
\label{tab4}
\end{table}

Table \ref{tab1} presents a comprehensive parameter study of $\alpha$ and $\sigma$, where we apply elastic transformations at both the image and label levels using PSPNet on the ISPRS Vaihingen dataset. 
    The results are measured in terms of mean Intersection-over-Union (mIoU), with the baseline (no augmentation used) yielding an mIoU of 77.41.
        When the transformation is applied at the `image' level, we observe a significant performance degradation. 
            Especially, with $\sigma$ fixed at 1, increasing $\alpha$ from 1 to 100 leads to a sharp drop in mIoU from 76.87 to 14.40.
                We conjecture that it is because, as highlighted in Fig. \ref{fig1}(b), object boundaries and textures are heavily distorted, often to the point of losing semantic consistency.
                    This type of warping may introduce non-realistic visual cues, which are harmful for training deep models that rely on spatial and textural continuity for accurate segmentation.
    In contrast, when the transformation is applied at the `label' level, the performance remains stable or even improves for moderate values of $\alpha$ and $\sigma$. 
        For example, at $\alpha$=15 and $\sigma$=5, the model achieves an mIoU of 77.70, surpassing the baseline. 
            As shown in Fig. \ref{fig1}(c), in this case, the input image remains visually intact, while the label undergoes spatial variation that helps the model generalize better to minor misalignments or boundary uncertainty.

    Moreover, to validate our belief that diversifying the combinations of $\alpha$ and $\sigma$ beyond the conventional elastic transform could lead to greater robustness, we conduct experiments by randomly sampling ($\alpha$, $\sigma$) pairs at each iteration during training (excluding all configurations where $\sigma$=1, due to its tendency to generate unrealistic, fragmented label masks).
            This setting is reported as ``All ($\alpha$, $\sigma$) combinations except for $\sigma$=1'' in Table \ref{tab1}.
                For the `image' level, it yields an mIoU of 70.69, still significantly below the baseline, emphasizing the limited benefit of image-level deformation. 
        However, for the `label' level (i.e., our \textbf{\textit{NSegment}}), the model achieves the best result of 77.75, indicating that label-level elastic deformation not only preserves semantic consistency but also enhances robustness through increased diversity. 
            As seen in Fig. \ref{fig3}, certain ($\alpha$, $\sigma$) combinations—such as (100, 3)—can introduce quite strong distortion to the segmentation label. 
                    Our method intentionally samples from a wide range of deformation strengths (as indicated in line 5 of Algorithm \ref{algo1}), spanning from mild to strong settings. 
                This stochastic sampling strategy exposes the model to diverse geometric variations of ground truth during training, thereby encouraging it to learn label-deformation-invariant representations, whereas simplifying the parameter search burden.

We previously showed the effectiveness of our proposed strategy, \textbf{\textit{NSegment}}, which dynamically applies distinct elastic transformations to each sample's label at every epoch.
    In Table \ref{tab2}, we additionally present a comparison between the conventional elastic transform—where identical deformations are applied to both the image and its label (as shown in Fig. \ref{fig1}(b))—and our approach, under randomly sampled ($\alpha$, $\sigma$) pairs\footnote{This setting was determined via preliminary grid search and empirical analysis. Alternative combinations for better performance may exist.}, \textcolor{black}{the Cartesian Product of two sets \{1, 15, 30, 50, 100\} and \{3, 5, 10\}}.
        While the naive elastic deformation (i.e., (a) in Table \ref{tab2}) method degrades performance by disrupting the visual realism of the input, \textbf{\textit{NSegment}} leverages `label-only elastic deformation' that enhances generalization without compromising semantic consistency (67.58 $\rightarrow$ 77.75).


\begin{figure*}[!ht]
    \centering
    \includegraphics[width=16cm]{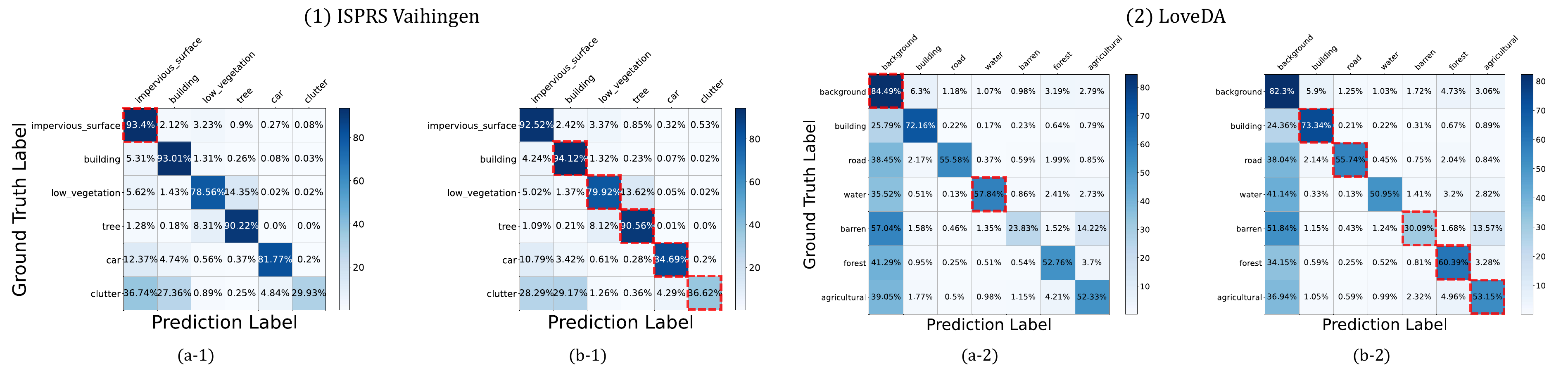} \vspace{-0.3cm}
    \caption{Normalized confusion matrices of the prediction results on the (1) ISPRS Vaihingen and (2) LoveDA datasets: (a) best-performing baseline results and (b) best-performing results of the proposed \textbf{\textit{NSegment}} method among five runs on (1) and (2) using DeepLabV3+. Each \textcolor{red}{\textbf{red}} dashed box indicates higher class-wise accuracy between (a) and (b) for each dataset. Best viewed on a colored screen with zoom.}
    \label{fig5}
    \vspace{-0.3cm}
\end{figure*}

\vspace{-0.2cm}
\subsection{Impact of \textbf{\textit{NSegment}} on existing RS segmentation models}
\label{sec3D}
This subsection demonstrates that the proposed \textbf{\textit{NSegment}} can effectively improve the performance of existing segmentation models.
For each dataset–model combination, we repeated the experiment 5 times with different random seeds (starting from the same weight initialization for 80,000 iterations).
    To provide a more objective assessment, we report all results as `\textbf{\textit{maximum (mean $\pm$ standard deviation)}}' of mIoU over five test-time checkpoints. This format is used consistently throughout Tables \ref{tab3}, \ref{tab4}, and \ref{tab5}. 
        We set $p$ to \textcolor{black}{$0.5$} and define $\mathbf{\Omega}=\{(\alpha_k, \sigma_k ) \}_k$ to \textcolor{black}{the Cartesian Product of two sets \{1, 15, 30, 50, 100\} and \{3, 5, 10\}} for our method.
    To demonstrate the effectiveness of our method, we adopt several state-of-the-art models\footnote{Section F of the supplementary material details models' configurations. } presented
    from 2017 to 2025, applicable to RS image segmentation and code available, as summarized in Table \ref{tab3}.
        Each baseline is trained without any data augmentation.
            

Notably, the results in Table \ref{tab3} demonstrate the robustness and generalizability of \textbf{\textit{NSegment}} across diverse datasets and model architectures.
     On the Vaihingen dataset—the smallest in our study (0.47 GB)—our method achieves the most substantial performance improvements (max: +0.80, mean: +0.38). 
        This suggests that our method might be particularly effective in small-data regimes, where overfitting and insufficient generalization can hinder the performance of baseline models.
    Interestingly, on the LoveDA dataset—the largest in our study (8.93 GB)—our method still yields noticeable improvements, with a peak of +0.62 and an average gain of +0.23.
        Despite its large size, LoveDA poses unique challenges due to its coarse spatial resolution and background diversity (urban/rural).

\textbf{Limitations}. In contrast, the performance gains on the Potsdam dataset (2.22 GB) are relatively modest (max: +0.26, mean: +0.09), leaving limited room for improvement.
    To further investigate this observation, we conduct an in-depth analysis across all datasets.
        From our analysis\footnote{See Section G of the supplementary material for detailed data analysis.}, we found that Potsdam contains a noticeably higher proportion of tiny (less than 10 pixels) segmentation masks compared to other datasets, where label-level deformations might lead to a greater risk of semantic misalignment.
    We plan to explore `scale-aware' perturbations in future work to mitigate this.
 Besides, as shown in Fig. \ref{fig5}, while the proposed method enhances the overall accuracy, the occasional accuracy decrease for a few classes suggests potential areas for further refinement. 

\vspace{-0.3cm}
\subsection{Combining \textbf{\textit{NSegment}} and existing augmentation methods}

To further investigate the applicability of \textbf{\textit{NSegment}}, we extend our experiments by incorporating several existing data augmentation techniques during training. 
    We consider `Horizontal Flipping' (HF) and `Random Resize' (RR) for our experiments.
        In light of the renewed interest in patch-based augmentations \cite{oh2024provable}, CutMix \cite{yun2019cutmix} (CM) is also integrated into our investigation.
            Each augmentation technique is activated with a probability of 0.5.
        Due to time and space constraints, this subsection solely adopts DeepLabV3+, while retaining the same experimental settings introduced in Section \ref{sec3D}.
As shown in Table \ref{tab4}, integrating \textbf{\textit{NSegment}} consistently improves segmentation performance across all datasets, regardless of the accompanying augmentation methods.
    For instance, applying ours with HF and RR (B + HF + RR + Ours) yields improvements over using only HF and RR (B + HF + RR) by +0.20, +0.21, and +3.99 in average on Vaihingen, Potsdam, and LoveDA, respectively.
Similar trends can be observed across all datasets when using ours with CM.
    The experimental results validate that \textbf{\textit{NSegment}} consistently improves model robustness and generalization ability, and can be effectively integrated alongside various existing augmentation strategies to achieve superior RS segmentation performance.
    
\vspace{-0.3cm}
\subsection{\textcolor{black}{Comparing \textbf{NSegment} and existing LNL methods}}

\begin{table}[t!]
\vspace{-0.1cm} 
\centering
\caption{{\label{tab:table-name} \textcolor{black}{ Comparing \textbf{\textit{NSegment}} with existing LNL methods, using DeepLabV3+ on the ISPRS Vaihingen dataset } }} \vspace{-0.2cm}
\resizebox{8.0cm}{!}{%
\begin{tabular}{@{}l|ccc@{}}
\toprule
\multicolumn{1}{c|}{\multirow{2}{*}{\begin{tabular}[c]{@{}c@{}}Each model is trained\\from the same weight initialization\end{tabular}}} & \multicolumn{1}{c}{\multirow{2}{*}{\begin{tabular}[c]{@{}c@{}}Test\\mIoU\end{tabular}}} & \multicolumn{1}{c}{\multirow{2}{*}{\begin{tabular}[c]{@{}c@{}}FLOPs\end{tabular}}} & \multicolumn{1}{c}{\multirow{2}{*}{\begin{tabular}[c]{@{}c@{}}Normalized Computational\\Complexity (FLOPs)\end{tabular}}} \\ \\ \midrule
Baseline (B)                                            & 77.98 (77.53$\pm$0.30) & \makebox[3em][r]{54.27}\,G & 1x \\
+ T. Kaiser \textit{et al.} \cite{kaiser2023compensation} (CVPR-W'23)            & 77.94 (77.68$\pm$0.30) & \makebox[3em][r]{55.04}\,G & 1.01x \\
+ L2B \cite{zhou2024l2b} (CVPR'24)                        & 78.00 (77.72$\pm$0.43) & \makebox[3em][r]{54.27}\,G & 1x \\
+ UCE \cite{landgraf2024uncertainty} (ISPRS-Annals'24)    & 78.01 (\underline{77.91}$\pm$0.12) & \makebox[3em][r]{542.69}\,G & 10x \\
+ AIO2 \cite{liu2024aio2} (TGRS'24)                       & \underline{78.02} (77.73$\pm$0.27) & \makebox[3em][r]{108.54}\,G & 2x \\
+ NBBOX \cite{kim2025nbbox} (GRSL'25)                     & 73.74 (73.39$\pm$0.30) & \makebox[3em][r]{54.27}\,G & 1x \\
\cellcolor[HTML]{FFCE93}+ Ours 
& \cellcolor[HTML]{FFCE93}\textbf{79.02} (\textbf{78.26}$\pm$0.58)  
& \cellcolor[HTML]{FFCE93}\makebox[3em][r]{54.27}\,G  
& \cellcolor[HTML]{FFCE93}1x \\ \bottomrule
\end{tabular}
} \vspace{-0.3cm}
\label{tab5}
\end{table}

\textcolor{black}{
    This subsection presents a comparative analysis between the \textbf{\textit{NSegment}} and existing `Leaning from Noisy Labels (LNL)\footnote{Section H of the supplementary material details training configurations of LNL approaches relevant to Table \ref{tab5}.}' methods that are code available.
    Given the limitations of time and space, we restrict our experiments to DeepLabV3+ with Vaihingen under the same settings as in Section \ref{sec3D}.
        While most LNL approaches have been studied for handling explicit annotation errors aside from NBBOX \cite{kim2025nbbox}, they often fall short in addressing the implicit label noise in RS image segmentation, as shown in Table \ref{tab5}.
    \textcolor{black}{Besides, some LNL methods entail high computational overhead due to intricate training pipelines or additional modules. }
        These observations highlight the practical utility and robustness of our \textbf{\textit{NSegment}} method in the context of inherent label noise in RS segmentation.
}

\vspace{-0.3cm}
\section{Conclusion}
In this work, we have proposed a novel data augmentation method, namely \textbf{\textit{NSegment}} for RS image segmentation.
    By applying elastic transformations exclusively to segmentation labels and varying the deformation intensity per training sample and epoch, our method mitigates the effects of annotation inconsistencies and implicit noises without increasing model complexity or training time.
        Our experimental results demonstrate that our method consistently improves segmentation performance across various state-of-the-art models.
    Notably, it can be easily integrated with other existing augmentation methods, making it a valuable tool for improving robustness and performance in real-world RS applications.
\vspace{-0.3cm}

\ifCLASSOPTIONcaptionsoff
  \newpage
\fi



%


\onecolumn
\appendix
\subsection{Comparing our \textbf{\textit{NSegment}} with \textbf{\textit{NBBOX}}}
\begin{figure}[!ht]
    \centering
    \includegraphics[width=18cm]{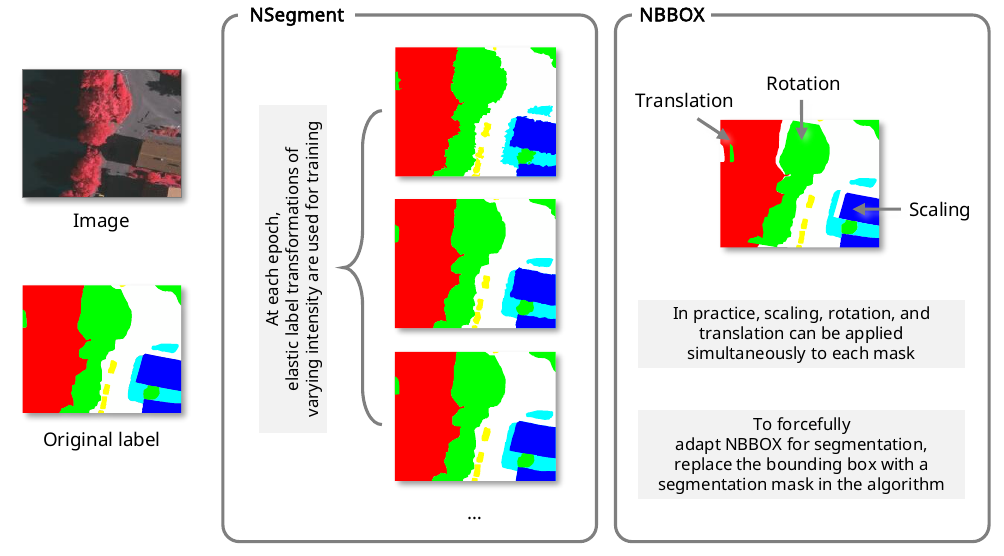} \vspace{-0.3cm}
    \caption{\textcolor{black}{Visual comparison between the proposed \textbf{\textit{NSegment}} for RS image segmentation and \textbf{\textit{NBBOX}} for RS object detection. }}
    \label{supfig1}
    \vspace{-0.0cm}
\end{figure}

The proposed \textbf{\textit{NSegment}} is inspired by \textbf{\textit{NBBOX}} \cite{kim2025nbbox}, a data augmentation technique initially developed for RS object detection. 
    \textbf{\textit{NBBOX}} has demonstrated the effectiveness of applying non-identical transformations to images and their associated bounding box labels. 
        Specifically, it injects noise only into oriented bounding boxes through random geometric perturbations (scaling, rotation, and translation), thereby enhancing model robustness against potential label inconsistencies.
            However, directly extending such rigid (or linear) transformations to RS image segmentation is non-trivial. 
    Unlike object detection, where the label is a compact geometric primitive (i.e., a bounding box), segmentation involves dense, pixel-level annotations that are highly sensitive to spatial misalignment.
        Hence, applying linear transforms (e.g., rotation or translation) solely to the segmentation mask can introduce severe semantic inconsistency between the input and its label as shown in Fig. \ref{supfig1}.

To overcome these limitations, \textbf{\textit{NSegment}} adopts elastic deformation as its core augmentation strategy, which is selectively applied only to the segmentation label. 
    This non-rigid transformation perturbs the label in a spatially smooth and locally stochastic manner, mimicking the soft and ambiguous boundary noise that frequently arises in real-world RS annotations due to shadows, occlusions, and mixed pixels. 
    It marks a key distinction between \textbf{\textit{NBBOX}} and \textbf{\textit{NSegment}}.
    



\vspace{-0.2cm}
\subsection{Source code availability}
In this study, we publicly release the implementation of \textbf{\textit{NSegment}} (\textbf{N}oise injection into \textbf{Segment}ation label) at \url{https://github.com/unique-chan/NSegment} to facilitate reproducibility and encourage further research.
    Detailed instructions for installation and usage are provided in the README.md file within the repository.
        Our proposed algorithm is implemented using Python 3, NumPy, OpenCV2, and MMSegmentation as follows:

\vspace{0.2cm}

\noindent\makebox[\textwidth] {
\centering
\begin{minipage}{18cm}
\label{supalgo1}
\begin{lstlisting}[label=algo1,caption=\small\textsc{Python implementation of the proposed method named \textbf{\textit{NSegment}}}.]
import cv2
import numpy as np

from mmseg.registry import TRANSFORMS


@TRANSFORMS.register_module()
class NoisySegment:
    def __init__(self, alpha_sigma_list=None, prob=0.5):
        if not alpha_sigma_list:
            alpha_sigma_list = [(alpha, sigma) for alpha in [1, 15, 30, 50, 100] 
                                 for sigma in [3, 5, 10]]
        self.alpha_sigma_list = alpha_sigma_list
        self.prob = prob

    def __call__(self, results):
        if np.random.rand() > self.prob:
            return results
        segment = results['gt_seg_map']
        noisy_segment = self.transform(segment)
        results['gt_seg_map'] = noisy_segment
        return results

    def transform(self, segment, random_state=None):
        if random_state is None:
            random_state = np.random.RandomState(None)
        alpha, sigma = self.alpha_sigma_list[random_state.randint(len(self.alpha_sigma_list))]
        shape = segment.shape[:2]
        dx = alpha * (2 * random_state.rand(*shape) - 1)
        dy = alpha * (2 * random_state.rand(*shape) - 1)
        dx = cv2.GaussianBlur(dx, (0, 0), sigma)
        dy = cv2.GaussianBlur(dy, (0, 0), sigma)
        x, y = np.meshgrid(np.arange(shape[1]), np.arange(shape[0]))
        map_x = np.clip(x + dx, 0, shape[1] - 1).astype(np.float32)
        map_y = np.clip(y + dy, 0, shape[0] - 1).astype(np.float32)
        noisy_segment = cv2.remap(segment, map_x, map_y, interpolation=cv2.INTER_NEAREST)
        return noisy_segment
\end{lstlisting}
\end{minipage}
}

\vspace{-0.2cm}
\subsection{Experimental environments}
All experimental evaluations were conducted on a dedicated workstation with an AMD Ryzen 7 3700X processor, an NVIDIA RTX 3090 GPU, and the Ubuntu 22.04 LTS operating system.

\vspace{-0.2cm}
\subsection{Detailed descriptions of datasets for experimentation}
The Vaihingen and Potsdam datasets are high-resolution aerial image benchmarks for evaluating semantic labeling methods in urban environments.
    The Vaihingen dataset consists of 33 high-resolution true orthophoto (TOP) tiles with an average size of 2494$\times$2064, captured over a residential area in Vaihingen, Germany, with a spatial resolution of 9cm.
            We used 16 tiles for training and the remaining 17 tiles for testing on Vaihigen, following previous studies like \cite{ma2023docnet, ma2025logcan++}.
    The Potsdam dataset features 38 higher-resolution TOP tiles of 6000$\times$6000 pixels, covering a larger and more diverse urban area of Potsdam in Germany, with a spatial resolution of 5cm.
        We used 24 tiles for training and the remaining 14 tiles for testing on Potsdam, following previous studies like \cite{ma2023docnet, ma2025logcan++}.
    Both datasets provide pixel-level annotations for six classes, including impervious surfaces, buildings, low vegetation, trees, cars, and clutter (background class).
        As the proportion of the clutter is only about 1\% in both datasets \cite{ma2024sam}, we excluded the clutter category for both datasets in our experiments.
        Moreover, following the prior work like \cite{luo2024fsegnet}, 
        we cropped each image tile into 512$\times$512 sized patches with a stride of 256.
            For the Vaihingen dataset, we used 344 cropped images for training and 398 cropped images for testing.
            For the Potsdam dataset, we used 3456 cropped images for training and 2016 cropped images for testing.

The LoveDA dataset is a large-scale satellite semantic segmentation benchmark for land cover classification with a spatial resolution of 0.3m.
    It includes 5987 annotated remote sensing images across both urban and rural regions in China, covering seven semantic classes: background, building, road, water, barren, forest, and agriculture.
        Unlike Vaihigen and Potsdam, we did not exclude the background class for LoveDA in our experiments following other prior work like \cite{ma2024sam}.
            For LoveDA, we used 2522 images for training and 1669 images for testing. 
            Note that for convenience, instead of using an online evaluation server, we consistently used the official validation set solely for testing purposes across all experiments.
                We still used the entire original training dataset as-is for training purposes.
            \textcolor{black}{For the sake of training convenience, each image is cropped to a size of 512$\times$512 during the training phase. To ensure a fair comparison and assess whether the integration of \textbf{\textit{NSegment}} consistently improves performance over the baseline models, we fixed the seed of the random module used for cropping.}
                

\vspace{-0.2cm}
\subsection{Understanding the roles of parameters $\alpha$ and $\sigma$ in \textbf{\textit{NSegment}}}
\begin{figure}[!ht]
    \centering
    \includegraphics[width=18cm]{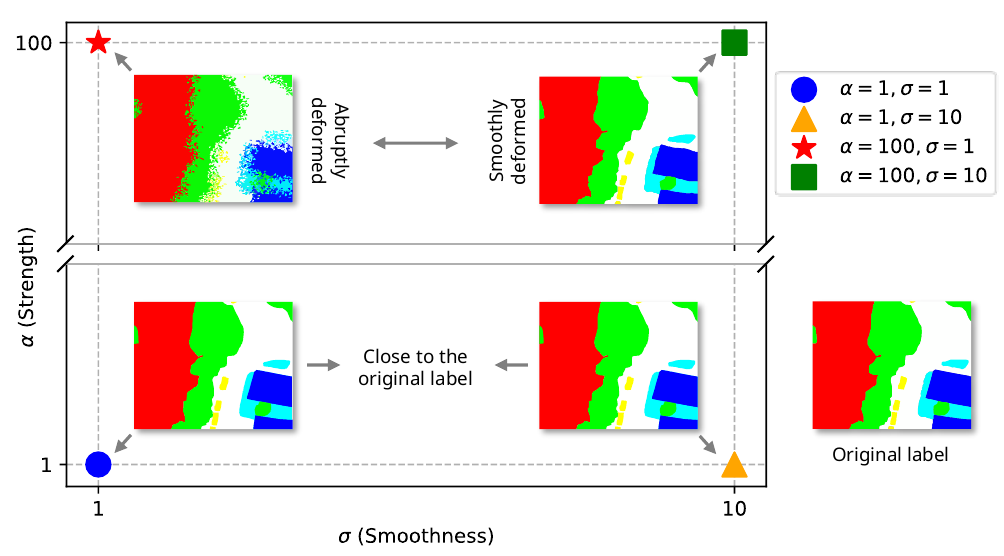} \vspace{-0.3cm}
    \caption{\textcolor{black}{Visualization of label deformation effects across four representative ($\alpha, \sigma$) configurations in \textbf{\textit{NSegment}}.}}
    \label{supfig2}
    \vspace{-0.0cm}
\end{figure}

We provide a focused discussion on the functional roles of the two key hyperparameters, $ \alpha$ and $\sigma$, used in our method.
    These parameters control the nature and intensity of the elastic deformation applied to segmentation label masks as follows:
\begin{itemize}
    \item $\alpha$ controls the magnitude (strength) of the displacement applied to each pixel in the label mask. 
        Higher values of $\alpha$ result in greater random displacement amplitudes, thereby producing more pronounced distortions in the label shape.
    Lower $\alpha$ values correspond to more conservative perturbations that resemble minor boundary noise.
    \item $\sigma$ determines the spatial coherence (smoothness) of the deformation field. It is the standard deviation in the Gaussian smoothing kernel.
        A small $\sigma$ leads to sharp, highly localized and irregular displacements, whereas larger $\sigma$ spreads the deformation smoothly over a broader area, preserving the structural continuity of the label.
\end{itemize}

\definecolor{deepgreen}{HTML}{008000}

We analyze four representative ($\alpha, \sigma$) configurations – (1, 1), (1, 10), (100, 1), and (100, 10) as shown in Fig. \ref{supfig2} to guide readers toward interpreting their functional roles.  
\begin{itemize}
    \item \textcolor{blue}{\ding{108}} ($\alpha=1, \sigma=1$) and \textcolor{orange}{\ding{115}} ($\alpha=1, \sigma=10$) – Minimal deformation: 
        When $\alpha$ is very small ($\alpha=1$), the displacement magnitude is minimal regardless of the choice of $\sigma$.
            Though $\sigma$ controls the spatial coherence of the deformation field, its influence becomes negligible when the underlying displacements are too small.
        Consequently, these settings produce outcomes similar to the original label.
    \item \textcolor{red}{\ding{72}} ($\alpha=100, \sigma=1$) – Strong and abrupt perturbation: When a large $\alpha$ is paired with a very small $\sigma$, the displacement becomes not only strong but also spatially erratic.
        The lack of smoothing causes the deformation to appear noisy and fragmented, leading to label masks that no longer resemble meaningful annotations.
    \item \textcolor{deepgreen}{\ding{110}} ($\alpha=100, \sigma=10$) – Moderate deformation: This setting results in a large but spatially smooth displacement field. Though the label structure is warped, the continuity of object boundaries is preserved due to the quite large smoothing factor $\sigma$ ($\sigma=10$). This kind of transformation simulates realistic annotation ambiguity and encourages the model to learn more invariant features.
\end{itemize}

Note that, rather than requiring manual tuning of exact $(\alpha, \sigma)$, \textbf{\textit{NSegment}} randomly samples from a wide set of such combinations, the Cartesian Product of two sets \{1, 15, 30, 50, 100\} and \{3, 5, 10\} at training time, which allows the model to see diverse perturbations without any sensitive hyperparamter calibration.
        As evidenced by the preceding examples, we exclude the case where $\sigma = 1$, as a moderate level of spatial smoothing is essential for ensuring the structural stability of labels.
    This design decision substantially reduces the practical burden of tuning while enhancing model generalization.


\subsection{Architectural and training details of the segmentation models used in this work}

Please refer to Tables \ref{suptab1} and \ref{suptab2}.

\begin{table*}[h!]
\vspace{-0.3cm} 
\centering
\caption{{\label{tab:table-name} Architectural details of state-of-the-art RS segmentation models used in this work.}} \vspace{-0.2cm}
\resizebox{18cm}{!}{%
\begin{tabular}{@{}ccccccccccc@{}}
\toprule
Model & PSPNet \cite{zhao2017pyramid} & DeepLabV3+ \cite{chen2018encoder} & DANet \cite{fu2019dual} & OCRNet \cite{yuan2020object} & SegFormer \cite{xie2021segformer} & DDRNet \cite{pan2022deep} & DOCNet \cite{ma2023docnet} & RS3Mamba \cite{ma2024rs} & LOGCAN++ \cite{ma2025logcan++} \\ \midrule
Backbone & ResNet-18 & ResNet-18 & ResNet-50 & HRNet-W18 & MiT-B0 & 23-Slim\_2xb6 & HRNet-W32 & VMamba\_Tiny & RepViT\_M2\_3   \\
\bottomrule
\end{tabular}
} \vspace{-0.2cm}
\label{suptab1}
\end{table*}

\begin{table*}[h!]
\vspace{-0.3cm} 
\centering
\caption{{\label{tab:table-name} Training details of state-of-the-art RS segmentation models used in this work for each dataset.}} \vspace{-0.2cm}
\resizebox{18cm}{!}{%
\begin{tabular}{@{}ccccc@{}}
\toprule
Model & Dataset & Training iterations & Optimizer & Learning rate scheduling \\ \midrule
\multirow{3}{*}{\makecell{PSPNet \cite{zhao2017pyramid}, DeepLabV3+ \cite{chen2018encoder}, DANet \cite{fu2019dual},\\ OCRNet \cite{yuan2020object}, SegFormer \cite{xie2021segformer}, DDRNet \cite{pan2022deep}}}   & Vaihingen &  &  &  \\
 & Potsdam & 80,000 & SGD(lr=0.01, momentum=0.9, weight\_decay=0.0005) & PolyLR(eta\_min=1e-4, power=0.9) \\
 & LoveDA &  &  &  \\ \midrule
\multirow{3}{*}{RS3Mamba \cite{ma2024rs}} & Vaihingen &  &  &  \\
 & Potsdam & 80,000 & SGD(lr=0.01, momentum=0.9, weight\_decay=0.0005) & MultiStepLR(milestones(epoch)=[25, 35, 45], gamma=0.1) \\
 & LoveDA &  &  &  \\ \midrule
\multirow{3}{*}{DOCNet \cite{ma2023docnet}, LOGCAN++ \cite{ma2025logcan++}} & Vaihingen &  &  &  \\
 & Potsdam & 80,000 & AdamW(lr=0.0001, momentum=0.9, weight\_decay=0.0001)& PolyLR(eta\_min=0, power=0.9) \\
 & LoveDA &  &  &  \\  \bottomrule
\end{tabular}
} \vspace{-0.2cm}
\label{suptab2}
\end{table*}

\subsection{Dataset-specific analysis of label size distributions and its impact on \textbf{\textit{NSegment}} effectiveness}
\begin{figure}[!ht]
    \centering
    \includegraphics[width=18cm]{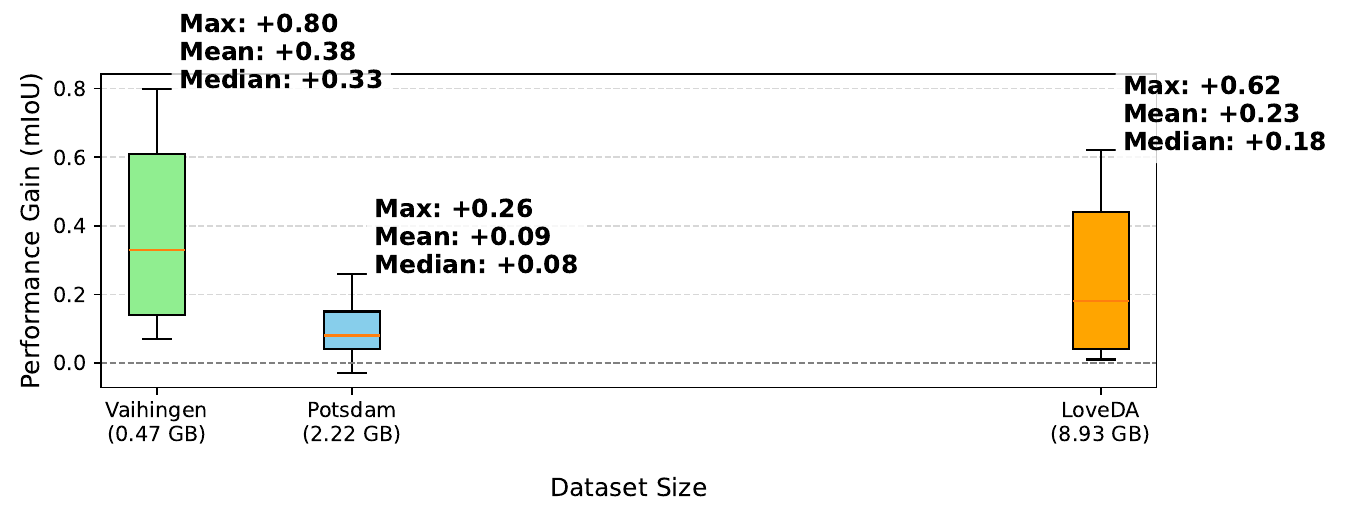} \vspace{-0.3cm}
    \caption{Performance gains (measured as the average of model-wise mIoU values) achieved by our \textbf{\textit{NSegment}} compared to baseline models on the ISPRS Vaihingen, Potsdam, and LoveDA datasets. Each boxplot summarizes the distribution of average performance improvements across various models.}
    \label{supfig3}
    \vspace{-0.1cm}
\end{figure}

As summarized in Fig. \ref{supfig3}, the effectiveness of \textbf{\textit{NSegment}} shows noticeable variation across ISPRS Vaihingen, Potsdam, and LoveDA datasets.
    \textbf{\textit{NSegment}} achieves the most substantial performance gain on Vaihingen, which is the smallest in size.
In contrast, the performance gains on Potsdam are notably more limited.
    Besides, while LoveDA is large in volume and yields greater gains than Potsdam, its improvements still fall short of those on Vaihingen.
To further understand the varying degrees of effectiveness of our \textbf{\textit{NSegment}} method across different datasets, we conducted a detailed investigation into the size distribution of segmentation masks in the training and test splits of the three benchmark datasets.

\begin{figure}[!ht]
    \centering
    \includegraphics[width=18cm]{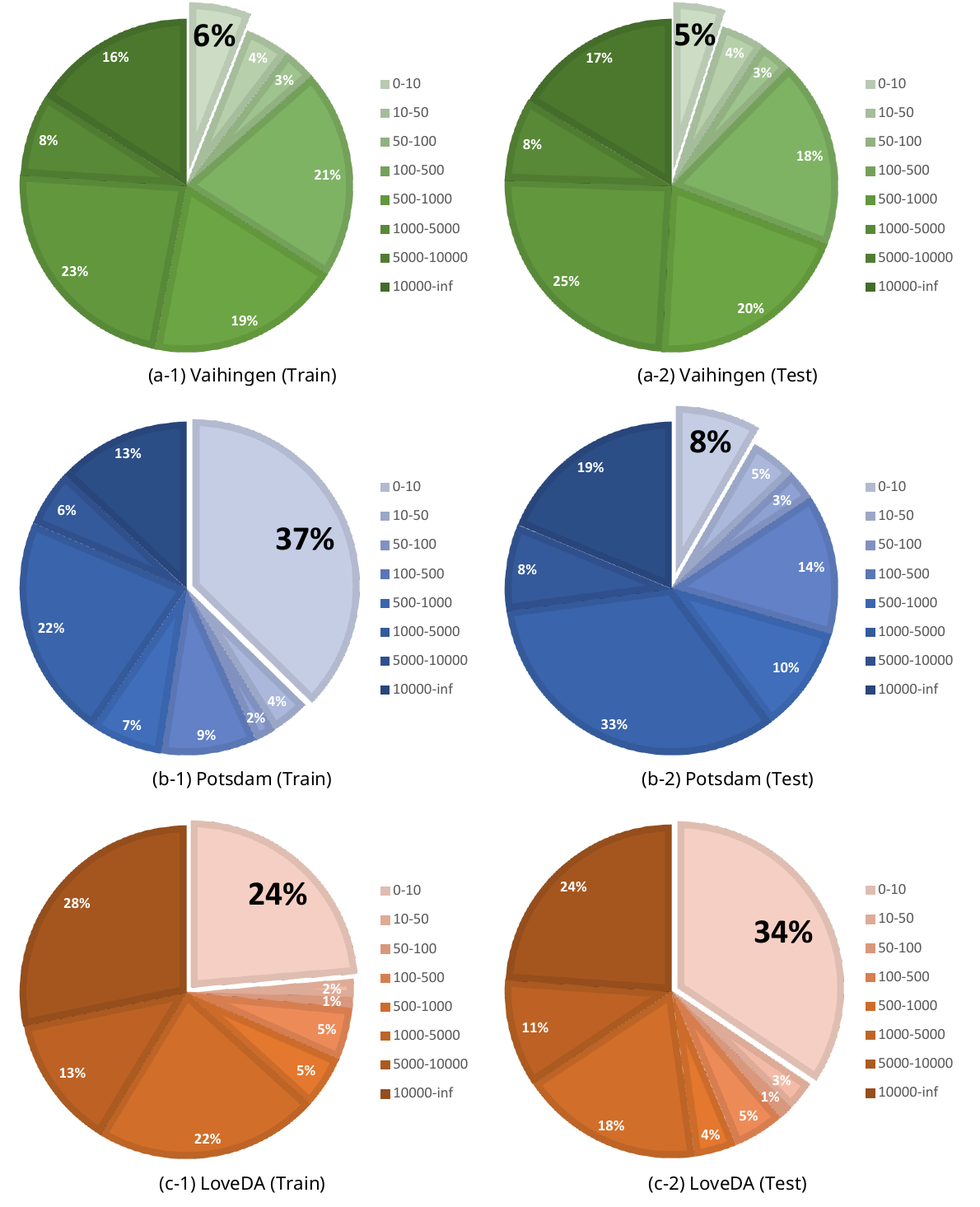} \vspace{-0.3cm}
    \caption{Area-based distributions of all segmentation masks in the training and test splits of (a) ISPRS Vaihingen, (b) ISPRS Potsdam, and (c) LoveDA datasets. Each pie chart shows the proportion of masks falling into specific area (pixels) intervals, aggregated across all classes. In particular, the proportion of tiny masks (with an area between 0 and 10 pixels) reaches 37\% in Potsdam (Train) and 24\% in LoveDA (Train), revealing a severe concentration of small-sized instances that may hinder robust model training. Note that, following prior works, we used the validation split as the test split for evaluation.}
    \label{supfig4}
    \vspace{-0.1cm}
\end{figure}

\begin{figure}[!ht]
    \centering
    \includegraphics[width=18cm]{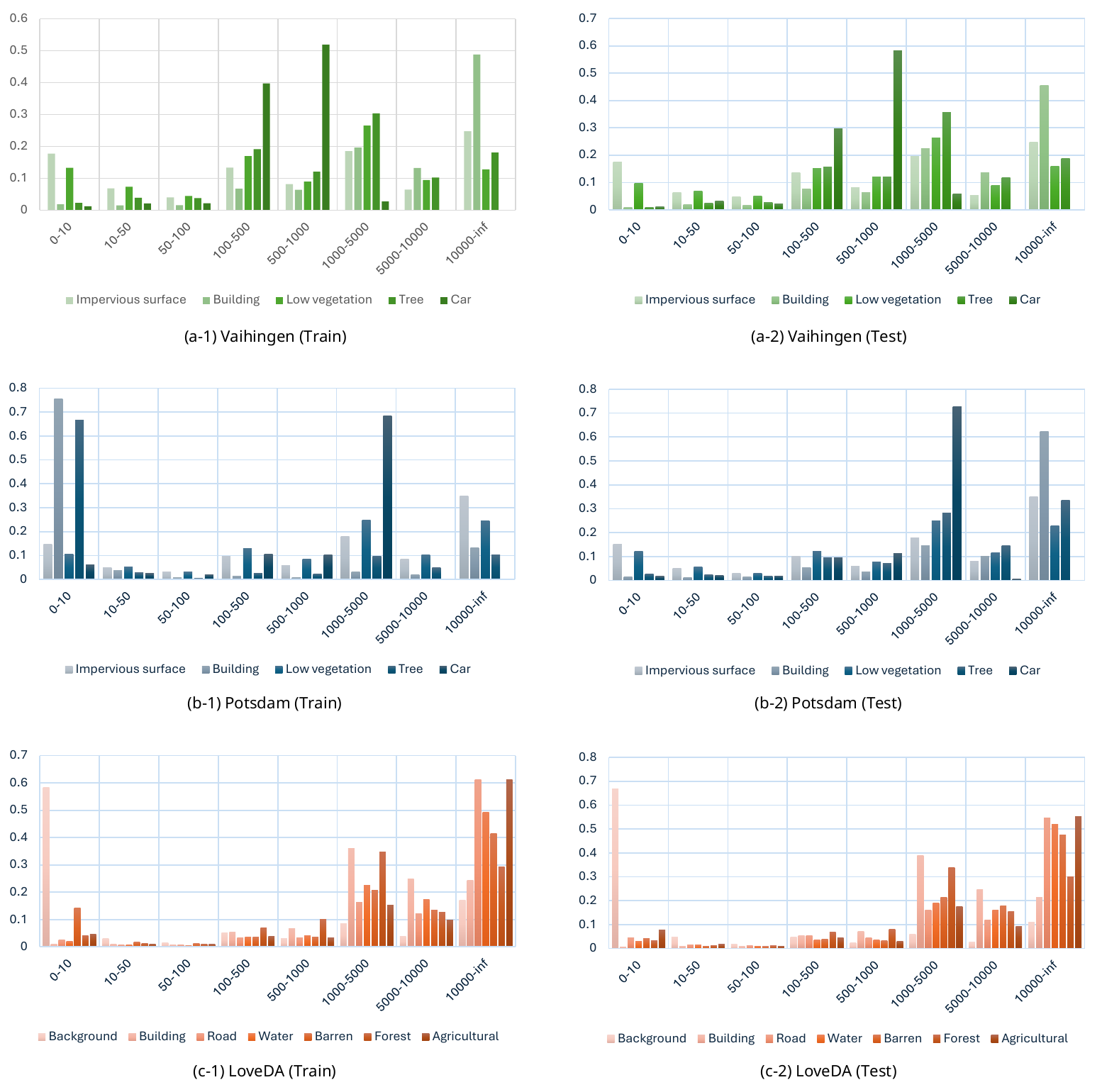} \vspace{-0.3cm}
    \caption{Distributions of class-wise mask sizes on training and test splits across three benchmark datasets: (a) ISPRS Vaihingen, (b) ISPRS Potsdam, and (c) LoveDA. For each dataset and its training/test partition, the $x$-axis indicates the discretized mask area (pixels) intervals, while the $y$-axis shows the proportion of class-wise masks whose area falls within each interval. Note that, as adopted in prior works, we utilized the validation split as the test set for performance evaluation, since no separate test annotations are provided for detailed analysis.}
    \label{supfig5}
    \vspace{-0.1cm}
\end{figure}

One of the clearest factors influencing the effectiveness of \textbf{\textit{NSegment}} is the proportion of tiny segmentation masks—defined as masks with area less than 10 pixels.
    Fig. \ref{supfig4} presents the distribution of segmentation mask areas (in pixels) for both the training and test splits of each dataset\footnote{Refer to Fig. \ref{supfig5} for the class-wise mask size distributions.}.
As visualized in Fig. \ref{supfig4}, the datasets vary significantly in this regard:
    \begin{itemize}
        \item \textbf{Vaihingen} maintains very low proportions in both splits (6\% and 5\%, respectively); \vspace{-0.0cm}
        \item \textbf{LoveDA} shows moderate levels in both splits (24\% in training, 34\% in testing);\vspace{-0.0cm}
        \item \textbf{Potsdam} exhibits the highest concentration of tiny masks in the training set (37\%), but only 8\% in the test set. \vspace{-0.0cm}
    \end{itemize}

These distributions align closely with the performance trends shown in Fig. \ref{supfig3}:
    \begin{itemize}
        \item \textbf{Vaihingen} yields the largest performance gain; \vspace{-0.0cm}
        \item \textbf{LoveDA} achieves a moderate gain; \vspace{-0.0cm}
        \item \textbf{Potsdam} shows the least improvement. \vspace{-0.0cm}
    \end{itemize}

This pattern suggests a \textbf{negative correlation between tiny object prevalence and our label-level augmentation benefit}, especially when such instances dominate training but are sparse at evaluation time (e.g., Potsdam).
        Small-scale instances are highly sensitive to even minor spatial perturbations. 
We conjecture that label-level elastic deformations in our \textbf{\textit{NSegment}} applied to these regions can easily lead to semantic misalignment, especially when deformation is not scale-aware.
    Hence, we plan to explore `scale-aware' perturbations in future work to mitigate this.

\subsection{\textcolor{black}{Training configurations of existing LNL methods} }

All methods were trained under identical settings: 80,000 iterations using SGD optimizer (initial\_learning\_rate= 0.01, momentum=0.9, weight decay=5e-4), and the learning rate was scheduled using a polynomial decay (PolyLR) with a power of 0.9 and a minimum learning rate of 1e-4.
    For each method, the detailed training configurations are as follows (Note that we aimed to preserve the original parameter settings reported in the respective papers as faithfully as possible.):
    \begin{itemize}
        \item \textbf{T. Kaiser et al.} \cite{kaiser2023compensation}: local\_compensation=True, loss\_balancing=0.01, non\_diagonal=True, symmetric=True, top\_k=5
        \item \textbf{L2B} \cite{zhou2024l2b}: cv2.erode and cv2.dilate are used for bootstrapping. For such operations, the kernel size is randomly chosen from \{7, 14, 21, 28, 35\} with probabilities \{0.1, 0.2, 0.5, 0.7, 1\}. Area thresholding(size=300) is applied using cv2.connectedComponents. The rotated angle is limited to the range of (-0.01, 0.01).
        \item \textbf{UCE} \cite{landgraf2024uncertainty}: num\_samples=10 (Each image is passed through the MC Dropout-enabled network 10 times to obtain stochastic predictions.)
        \item \textbf{AIO2} \cite{liu2024aio2}: (shift, erosion, dilation, rotation, remove) operations are applied. Area thresholding(size=30) is applied using cv2.connectedComponents. The rotated angle is limited to the range of (0.01, 1.0).
        \item \textbf{NBBOX} \cite{kim2025nbbox}: scaling\_range=(0.99, 1.01), rotation\_range=(-0.01, 0.01), translation\_range=(-1, 1), threshold=16
    \end{itemize}

%








\end{document}